# Horizontally Scalable Submodular Maximization

**Mario Lucic**[1]   LUCIC@INF.ETHZ.CH
**Olivier Bachem**[1]   OLIVIER.BACHEM@INF.ETHZ.CH
**Morteza Zadimoghaddam**[2]   ZADIM@GOOGLE.COM
**Andreas Krause**[1]   KRAUSEA@ETHZ.CH

[1]Department of Computer Science, ETH Zurich, Switzerland
[2]Google Research, New York

## Abstract

A variety of large-scale machine learning problems can be cast as instances of constrained submodular maximization. Existing approaches for distributed submodular maximization have a critical drawback: The *capacity* – number of instances that can fit in memory – must *grow* with the data set size. In practice, while one can provision many machines, the capacity of each machine is limited by physical constraints. We propose a truly scalable approach for distributed submodular maximization under fixed capacity. The proposed framework applies to a broad class of algorithms and constraints and provides theoretical guarantees on the approximation factor for any available capacity. We empirically evaluate the proposed algorithm on a variety of data sets and demonstrate that it achieves performance competitive with the centralized greedy solution.

## 1. Introduction

Submodularity is a powerful concept that captures the natural notion of diminishing returns. As such, many important machine learning problems can be formulated as constrained submodular maximization. Examples include influence maximization (Kempe et al., 2003), document summarization (Lin & Bilmes, 2011), and active learning (Golovin & Krause, 2011). Solving these optimization problems on a massive scale is a great challenge and trading off computational resources and the approximation quality becomes paramount.

Consequently, distributed computing has received a great deal of interest. One of the most desirable properties of distributed algorithms is *horizontal scaling* – scaling to larger instances by adding more machines of fixed capacity.

Prominent approaches in distributed submodular optimization are often based on a two-round divide and conquer strategy. In the first round, the data is partitioned so that each partition can fit on one machine. Then, the subproblems are solved for each partition in parallel. In the second round, the partial solutions are collected on one machine and used to calculate the final solution. It was recently shown that this class of algorithms enjoys good empirical and theoretical performance, often matching those of the centralized solution (Mirzasoleiman et al., 2013).

However, existing algorithms for distributed submodular maximization are not *truly* horizontally scalable. In fact, they rely on the implicit assumption that the capacity of each machine grows with the data set size. Consider the submodular maximization problem with a large ground set of size $n$ and cardinality constraint $k$. Existing approaches perform well in distributing computation in the initial round. However, problems materialize when the size of the union of all solutions computed across the cluster exceeds the capacity of a single machine. At that point, one cannot scale the existing approaches by adding more machines and they will simply break down. For the popular GREEDI algorithm, this occurs if the capacity is less than $\mathcal{O}(\sqrt{nk})$ (Mirzasoleiman et al., 2013). Since this will invariably happen as $n$ grows, this aspect is critical for true horizontal scaling and cannot be neglected.

**Our contribution.** We investigate and address the problems arising from fixed machine capacity. To the best of our knowledge, we present the first *horizontally scalable* framework for constrained submodular maximization. For a broad class of underlying algorithms, it enables distributed submodular maximization under cardinality constraints. For the popular greedy algorithm, the results extend to hereditary constraints. We empirically evaluate the proposed approach and demonstrate that it achieves performance close to the centralized greedy solution.

## 2. Background and Related Work

Let $V$ be a ground set with cardinality $n$. Consider a set function $f : 2^V \to \mathbb{R}^+ \cup \{0\}$. Function $f$ is monotone iff for any two subsets $X \subseteq Y \subseteq V$, $f(X) \leq f(Y)$. Function $f$ is submodular if and only if for any two subsets $X \subseteq Y \subseteq V$, and an item $e \in V \setminus Y$ it holds that

$$f(X \cup \{e\}) - f(X) \geq f(Y \cup \{e\}) - f(Y).$$

We focus on the problem of maximizing a non-negative monotone submodular function subject to a cardinality constraint. Given an integer $k$, the maximizer of $f$ is

$$\text{OPT} = \arg\max_{S \subseteq V : |S| \leq k} f(S).$$

The classic GREEDY algorithm (Nemhauser et al., 1978) which starts with an empty set and iteratively adds items with highest marginal gain achieves a $(1 - 1/e)$-approximation. As the algorithm is not feasible for massive datasets, our goal is to find OPT using a distributed algorithm with value oracle access to $f$.

Mirzasoleiman et al. (2013) developed the first two-round algorithm whereby in the first round the data is arbitrarily partitioned to $m$ machines and each machine runs GREEDY to select at most $k$ items. The partial solutions are then placed on a single machine for another run of GREEDY. This approach guarantees an approximation factor of $1/\Theta(\min(\sqrt{k}, m))$. Recently, Barbosa et al. (2015a) showed that instead of assuming an arbitrary partitioning of the data, one obtains an approximation factor of $(1-1/e)/2$ if the data is randomly partitioned. Mirrokni & Zadimoghaddam (2015) prove that GREEDY computes representative subsets – *coresets*, and that by selecting $\mathcal{O}(k)$ instead of $k$ items on each machine in the first round one obtains an approximation factor of $0.545$.

The implicit assumption in these algorithms is that that the capacity per machine $\mu \geq \max(n/m, mk)$, which implies $\mu \geq \sqrt{nk}$, optimizing over $m$. Hence, the capacity needs to grow with the data set size! As a result, they are not truly horizontally scalable, as the maximum available memory on each machine is essentially fixed.

Kumar et al. (2013) provide two multi-round algorithms for cardinality constrained submodular maximization. The THRESHOLDMR algorithm provides a $(1/2-\varepsilon)$-approximation with high probability, in $\mathcal{O}(1/\delta)$ rounds and using $\mathcal{O}(kn^\delta \log n)$ memory per machine. Furthermore, a variant of the algorithm, GREEDYSCALING, provides a $(1-1/e)/(1+\varepsilon)$ approximation in $\mathcal{O}((\log \Delta)/\varepsilon\delta)$ rounds using $\mathcal{O}(n \log(n)/\mu)$ machines with $\mathcal{O}(kn^\delta \log n)$ capacity per machine, with high probability. There are four critical differences with respect to our work. Firstly, each call to the pruning procedure requires a capacity of $\sqrt{2nk} \log n$, with high probability. However, given this capacity our algorithm always terminates in two rounds and we empirically observe approximation ratios very close to one. Secondly, our algorithm requires capacity of $kn^\delta$ to complete in $1/\delta$ rounds, while their result requires capacity of $kn^\delta \log n$. As a result, we require capacity greater than $k$, while they require capacity of at least $k \log n$ which grows with $n$. In the case of GREEDYSCALING the number of rounds is even greater than $\mathcal{O}(1/\delta)$, especially for a small $\varepsilon$ or a large $\Delta$. Thirdly, for the THRESHOLDMR algorithm one needs to try $\log(\Delta)/\epsilon$ thresholds which necessitates $n(\log \Delta)/(\mu\epsilon)$ machines compared to our $n/\mu$ machines, which is optimal and critical in practice where small $\varepsilon$ is desired. Finally, the pruning subprocedure that the aforementioned approach relies on has to be monotone, which is not the case for both the classic greedy and stochastic greedy. In contrast, we can use their thresholding-based algorithm as a compression subprocedure.

Barbosa et al. (2015b) proved that for any sequential algorithm that achieves an approximation factor of $\alpha$ and is *consistent*, there exists a randomized distributed algorithm that achieves a $(\alpha - \varepsilon)$-approximation with constant probability in $\mathcal{O}(1/\varepsilon)$ rounds. In particular, one can achieve a $(1 - 1/e - \varepsilon)$-approximation in $\mathcal{O}(1/\varepsilon)$ rounds and $\Theta(k\sqrt{nk}/\varepsilon^2)$ capacity per machine.

| ALGORITHM | MIN. CAPACITY | ROUNDS | ORACLE EVALUATIONS | APPROXIMATION |
|---|---|---|---|---|
| GREEDI | $\sqrt{nk}$ | 2 | $\mathcal{O}(nk)$ | $1/\Theta(\min(\sqrt{k}, m))$ |
| RANDGREEDI | $\sqrt{nk}$ | 2 | $\mathcal{O}(nk)$ | $(1-1/e)/2$ |
| RANDOMIZED CORESET | $\sqrt{nk}$ | 2 | $\mathcal{O}(nk)$ | $0.545$ |
| GREEDY SCALING | $\mathcal{O}(n^\delta k \log n)$ | $\mathcal{O}(\frac{1}{\varepsilon\delta} \log \Delta)$ | $\mathcal{O}(\frac{n}{\varepsilon} \log \Delta)$ | $1 - 1/e - \varepsilon$ |
| THRESHOLDMR | $\mathcal{O}(n^\delta k \log n)$ | $\mathcal{O}(\frac{1}{\delta})$ | $\mathcal{O}(\frac{n}{\varepsilon} \log k)$ | $1/2 - \varepsilon$ |
| DTHRESHGREEDY | $\Theta(\sqrt{nk}/\varepsilon)$ | $\mathcal{O}(1/\varepsilon)$ | $\mathcal{O}(nk^3/\varepsilon)$ | $1 - 1/e - \varepsilon$ |
| OUR RESULTS | $n$ | 1 | $\mathcal{O}(nk)$ | $1 - 1/e$ |
|  | $\sqrt{nk}$ | 2 | $\mathcal{O}(nk)$ | $(1-1/e)/2$ |
|  | $\mu > k$ | $r = \lceil \log_{\mu/k} n/\mu \rceil + 1$ | $\mathcal{O}(nk)$ | $1/2r$ |

Table 1. Distributed algorithms for submodular maximization with a cardinality constraint. The minimum capacity for existing algorithms grows with the ground set size. If necessary, the proposed algorithm trades off the capacity and the approximation factor.

Horizontally Scalable Submodular Maximization# 3. A Practical Multi-Round Framework

In contrast to the previous approaches, we focus on scalable constrained submodular maximization with fixed machine capacity. We will first focus on cardinality constrained optimization and later extend the results to optimization under hereditary constraints. To this end, we first formally define the problem and introduce the framework.

**Problem definition.** We are given oracle access to a monotone non-negative submodular function $f$ defined over a ground set $V$ of size $n$ and an integer $k$. Furthermore, we have access to a potentially unlimited set of machines $M$ where each machine has limited capacity $\mu$. Our goal is to compute a solution $S \subseteq V, |S| \leq k$ under these capacity constraints such that $f(S)$ is competitive with the optimal solution in the case of unlimited capacity.[1]

**Framework.** We propose the following multi-round framework which trades off machine capacity for approximation quality. The main idea is to maintain a set of items $A$ and keep compressing it until it fits on one machine. The compression is performed by a single machine algorithm $\mathcal{A}$ which receives a set of items as input and outputs a set of at most $k$ items.

In the first round, we set $A_0 = V$, provision $m_0 = \lceil n/\mu \rceil$ machines and randomly partition $A_0$ to those machines. To ensure that different parts of the random partition have almost equal size we perform the partitioning as follows. To partition $N$ items to $L$ parts, we assign each of the $L$ parts $\lceil N/L \rceil$ virtual free locations. We pick items one by one, and for each one we find a location uniformly at random among the available locations in all machines, and assign the item to the chosen location.

Each machine then runs algorithm $\mathcal{A}$ to select at most $k$ items from the items assigned to it. For the second round, we set $A_1$ to be the union of all $m_0$ partial solutions and provision $m_1 = \lceil |A_1|/\mu \rceil \leq \lceil km_0/\mu \rceil = \lceil kn/\mu^2 \rceil$ machines. The algorithm proceeds iteratively until $|A_t| \leq \mu$. We return the set with the maximum value among all partial solutions and the solution on $A_t$. Pseudo-code is provided in Algorithm 1 and one round is illustrated in Figure 1. We note that at this point the compression algorithm $\mathcal{A}$ is not yet specified.

To obtain the approximation guarantee of this framework we first upper bound the number of rounds and then the expected loss in each round. In this tree-based compression scheme, the number of rounds depends on the maximum branching factor allowed by the capacity. On the other hand, the expected loss depends on the probability of pruning elements from the optimal solution in each round.

---
[1] For the moment assume that $f$ can be computed without access to the full dataset (e.g. as in the active set selection problem).

**Algorithm 1** TREE-BASED COMPRESSION

1: **Input:** Set $V$, $\beta$-nice algorithm $\mathcal{A}$, $k$, capacity $\mu$.
2: **Output:** Set $S \subseteq V$ with $|S| \leq k$.
3: $S \leftarrow \emptyset$
4: $r \leftarrow \lceil \log_{\mu/k} n/\mu \rceil + 1$
5: $A_0 \leftarrow V$
6: **for** $t \leftarrow 0$ **to** $r-1$ **do**
7: $\quad m_t \leftarrow \lceil |A_t|/\mu \rceil$
8: $\quad$ Partition $A_t$ randomly into $m_t$ sets $T_1, \ldots, T_{m_t}$.
9: $\quad$ **for** $i \leftarrow 1$ **to** $m_t$ **in parallel do**
10: $\quad\quad S_i \leftarrow \mathcal{A}(T_i)$
11: $\quad\quad$ **if** $f(S_i) > f(S)$ **then**
12: $\quad\quad\quad S \leftarrow S_i$
13: $\quad A_{t+1} \leftarrow \cup_{i=1}^{m_t} S_i$
14: **return** $S$

**Proposition 3.1.** *Let $n \geq \mu > k$. Then, the number of rounds performed by Algorithm 1 is bounded by*

$$r \leq \lceil \ln(\mu/n)/\ln(k/\mu) \rceil + 1 = \lceil \log_{\mu/k} n/\mu \rceil + 1.$$

The result follows from the fact that in each round we reduce the size of $A$ by a factor of $\mu/k$.

To instantiate the framework, one needs to provide the compression algorithm $\mathcal{A}$ that will be run in each round on each machine in parallel. A natural choice is the class of $\beta$-nice algorithms (Mirrokni & Zadimoghaddam, 2015) which was introduced in the context of randomized composable coresets for submodular optimization.

**Definition 3.2.** *(Mirrokni & Zadimoghaddam, 2015) Consider a submodular set function $f$. Let $\mathcal{A}$ be an algorithm that given any $T \subseteq V$ returns subset $\mathcal{A}(T) \subseteq T$ with size at most $k$. Algorithm $\mathcal{A}$ is a $\beta$-nice algorithm for function $f$ and some parameter $\beta$ iff for any set $T \subseteq V$ and any item $x \in T \setminus \mathcal{A}(T)$ the following two properties hold:*

$$\mathcal{A}(T \setminus \{x\}) = \mathcal{A}(T) \quad (1)$$
$$f(\mathcal{A}(T) \cup \{x\}) - f(\mathcal{A}(T)) \leq \beta f(\mathcal{A}(T))/k \quad (2)$$

Intuitively, (1) ensures that the output of the algorithm does not depend on the items it did not select, and (2) guarantees that the marginal value of any item that was not selected is bounded by the average contribution of selected items. For instance, the classic GREEDY with consistent tie-breaking is 1-nice, and the THRESHOLDING GREEDY algorithm of Badanidiyuru & Vondrák (2014) is $(1 + 2\varepsilon)$-nice. The STOCHASTIC GREEDY algorithm (Mirzasoleiman et al., 2015) is a good choice for some problems. While the aforementioned algorithm has not been shown to be a $\beta$-nice, we demonstrate that it performs well empirically in Section 4.4

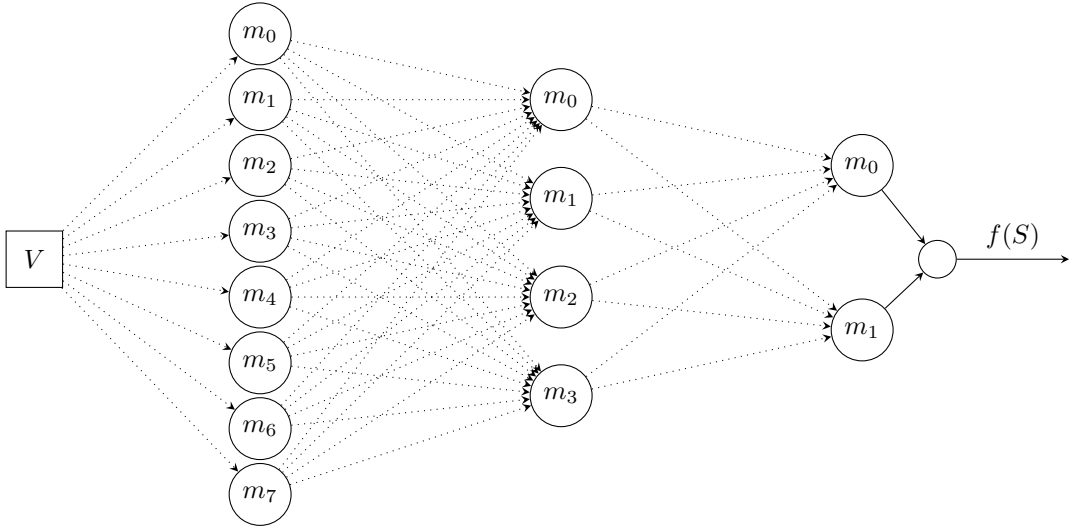

*Figure 1.* Illustration of distributed submodular maximization using Algorithm 1. The example displays a problem where $n = 16k$ and $\mu = 2k$. Initially, the ground set $V$ is randomly partitioned to 8 machines and solutions of size $k$ are generated using $\mathcal{A}$. The resulting $8k$ elements are again randomly partitioned in the second round. This continues until in the fourth round there are $2k$ elements left on a single machine at which point the final solution of $k$ elements is computed.

### 3.1. Approximation Factor for Cardinality Constraints

For a fixed $\beta$-nice algorithm, the approximation factor after multiple rounds depends only on the number of rounds and $\beta$. The following theorem relates the available capacity with the approximation guarantee of the proposed distributed multi-round framework.

**Theorem 3.3.** *Let $f$ be a monotone non-negative submodular function defined over the ground set $V$ of cardinality $n$, and $k$ the cardinality constraint. Let $\mathcal{A}$ in Algorithm 1 be a $\beta$-nice algorithm and $\mu$ the capacity of each machine. Then, Algorithm 1 yields a set $S$ of size at most $k$ with*

$$\mathbb{E}\big[f(S)\big] \geq \begin{cases} \frac{1}{1+\beta} f(\text{OPT}), & \text{if } \mu \geq n \\ \frac{1}{2(1+\beta)} f(\text{OPT}), & \text{if } n > \mu \geq \sqrt{nk} \\ \frac{1}{r \cdot (1+\beta)} f(\text{OPT}), & \text{otherwise,} \end{cases}$$

*with $r = \lceil \log_{\mu/k} n/\mu \rceil + 1$, using at most $\mathcal{O}(n/\mu)$ machines. If we use GREEDY as the $\beta$-nice algorithm, the approximation factor will be at least $(1 - 1/e)$ for $\mu \geq n$, $(1 - 1/e)/2$ for $\mu \geq \sqrt{nk}$, and $1/2r$ for arbitrary $r$.*

The first two cases are due to the classic analysis of GREEDY and the result of Barbosa et al. (2015a), respectively. We will focus on the third case in which the limited machine capacity gives rise to multiple rounds. To estimate the quality of the compression scheme, we will track how much of OPT is pruned in each round. Clearly, losing a constant fraction would lead to an exponential decrease of the approximation quality with respect to the number of rounds. A more promising approach is based on bounding the additive loss incurred in each round. The follow-ing Lemma is a generalization of a result from Mirrokni & Zadimoghaddam (2015) in that it holds for *any subset*, not only OPT. The proof is provided in the Section A of the supplementary materials.

**Lemma 3.4.** *Consider an arbitrary subset $B \subseteq V$, and a random partitioning of $B$ into $L$ sets $T_1, T_2, \cdots, T_L$. Let $S_i$ be the output of algorithm $\mathcal{A}$ on $T_i$. If $\mathcal{A}$ is $\beta$-nice, for any subset $C \subseteq B$ with size at most $k$, it holds that*

$$\mathbb{E}\big[f(C^S)\big] \geq f(C) - (1+\beta)\mathbb{E}\left[\max_{1 \leq i \leq L} f(S_i)\right]$$

*where $C^S = C \cap \left( \cup_{1 \leq i \leq L} S_i \right)$.*

The proof of Theorem 3.3 follows from an iterated application of Lemma 3.4 and a bound on the number of rounds for a fixed capacity.

**Proof of Theorem 3.3** Let $\text{OPT}^t$ be $\text{OPT} \cap A_t$ for $0 \leq t \leq r+1$. In particular, $\text{OPT}^0$ is OPT, and $\text{OPT}^{r+1}$ is the items of OPT that survive till the end of algorithm and are present in the last set that the only machine in round $r$ outputs. Since the output set $S$ of Algorithm 1 has the maximum value among the returned sets by all machines in all rounds,

$$f(S) \geq f(\text{OPT}^{r+1}).$$

To get the desired approximation factor, it suffices to bound the reduction in the value of remaining optimum items from round $t$ to $t+1$ for each $0 \leq t \leq r$. By applying Lemma 3.4, and setting $B = A_t$, $L = m_t$, and $C = \text{OPT}^t$, it follows that $\mathbb{E}[f(\text{OPT}^t) - f(\text{OPT}^{t+1})]$ is at most $1 + \beta$ times the maximum value of the $m_t$ output sets in round $t$, and consequently it is at most $(1+\beta)\mathbb{E}[f(S)]$ by definition of $S$. Since in the last round we have only one machine, we can get a better bound on $f(\text{OPT}^r) - f(\text{OPT}^{r+1})$ as follows. From the fact that $\mathcal{A}$ is a $\beta$-nice algorithm, it follows that

$$f(\text{OPT}^{r+1} \cup \{x\}) - f(\text{OPT}^{r+1}) \leq \beta f(\text{OPT}^{r+1})/k \quad (3)$$

for any $x \in \text{OPT}^r$. By submodularity of $f$, for any pair of sets $A$ and $B$, it holds that

$$\sum_{x \in A} f(B \cup \{x\}) - f(B) \geq f(A) - f(B).$$

In particular, summing over all $x \in \text{OPT}^r$ in Equation 3 yields $f(\text{OPT}^r) - f(\text{OPT}^{r+1}) \leq \beta f(\text{OPT}^{r+1})$, since there are at most $k$ items in $\text{OPT}^r$. We conclude that

$$\begin{aligned}
\mathbb{E}[f(S)] &\geq \mathbb{E}[f(\text{OPT}^{r+1})] \\
&\geq \mathbb{E}[f(\text{OPT}^r)]/(1+\beta) \\
&\geq (f(\text{OPT}) - (r-1)(1+\beta)\mathbb{E}[f(S)])/(1+\beta)
\end{aligned}$$

which implies

$$\mathbb{E}[f(S)] \geq \frac{1}{r \cdot (1+\beta)} f(\text{OPT}).$$

□

### 3.2. Approximation Factor for Hereditary Constraints

A constraint $\mathcal{I}$ is a family of feasible subsets of the ground set $V$. The constraint $\mathcal{I}$ is *hereditary* if for any $S \in \mathcal{I}$ all subsets of $S$ are also in $\mathcal{I}$. In submodular maximization under hereditary constraints the goal is to find a feasible set that maximizes $f$, i.e.,

$$\text{OPT} = \max_{S \in \mathcal{I}} f(S).$$

Examples include cardinality constraints $\mathcal{I} = \{A \subseteq V : |A| \leq k\}$, matroid constraints, where $\mathcal{I}$ corresponds to the collection of independent sets of the matroid, knapsack constraints, where $\mathcal{I} = \{A \subseteq V : \sum_{i \in A} w_i \leq b\}$, as well as arbitrary combinations of such constraints.

In this section, we prove that the proposed algorithm can be used for submodular maximization with any arbitrary hereditary constraint $\mathcal{I}$. In particular, if GREEDY is used as the compression procedure, the algorithm achieves approximation factor of $\mathcal{O}\left(\frac{\alpha}{r}\right)$, where $\alpha$ is the approximation factor of centralized GREEDY for submodular maximization under hereditary constraint $\mathcal{I}$. As such, the results in this section generalize Theorem 3.3.

The following theorem relates the available capacity with the approximation guarantee of the distributed multi-round framework using algorithm GREEDY as the $\beta$-nice algorithm for any hereditary constraint.

**Theorem 3.5.** *Let $f$ be a monotone non-negative submodular function defined over the ground set $V$ of cardinality $n$, and $\mathcal{I}$ the hereditary constraint. Let $\mathcal{A}$ in Algorithm 1 be GREEDY and $\mu$ the capacity of each machine such that $n \geq \mu > k$. Then, Algorithm 1 yields a set $S \in \mathcal{I}$ with*

$$\mathbb{E}[f(S)] \geq \frac{\alpha}{r} f(\text{OPT})$$

*with $r = \lceil \log_{\mu/k} n/\mu \rceil + 1$, using at most $\mathcal{O}(n/\mu)$ machines where $\alpha$ is the approximation factor of GREEDY for maximizing $f$ with constraint $\mathcal{I}$ on a single machine.*

The proof builds upon the rather elegant technique from Barbosa et al. (2015a). We start with defining the Lovász extension of submodular function $f$. For any vector $v = \{v_1, v_2, \cdots, v_n\}$, the Lovász extension is defined as

$$f^L(v) = \mathbb{E}_{\tau \sim \mathcal{U}[0,1]}[f(\{i \mid v_i \geq \tau\})].$$

We note that $\tau$ is drawn from the uniform distribution on the interval $[0,1]$. Submodularity of $f$ implies the following three properties on $f^L$:

(A) $f^L(c \cdot v) \geq c \cdot f^L(v)$ for any $0 \leq c \leq 1$.

(B) Function $f^L$ is convex.

(C) For any $S \subseteq V$, $f^L(\mathbb{1}_S) = f(S)$ where $\mathbb{1}_S$ is the vector with 1 at the $i$'th entry if $i \in S$, and 0 otherwise.

We now have all the prerequisites to prove the main result. Intuitively, our goal is to lower-bound the expected value of the selected elements by a fraction of the expected value of the pruned elements in each round.

**Proof of Theorem 3.5** Denote the output of GREEDY on set $A$ by GREEDY$(A)$. For any round $1 \leq t \leq r$, and machine index $1 \leq i \leq m_t$, let $T_i^t$ be the set of items sent to machine $i$ at round $t$. For any item $x \in V$, and round $1 \leq t \leq r$, let $p_x^t$ be the probability that item $x$ is not selected by GREEDY if it is sent to a random machine in round $t$. In other words,

$$p_x^t = \mathcal{P}(x \notin \text{GREEDY}(T_i^t \cup \{x\})),$$

for $i$ is chosen uniformly at random from $\{1, 2, \ldots, m_t\}$. Let $O^t$ be the set of these omitted items, i.e.

$$O^t = \{x \mid x \notin \text{GREEDY}(T_i^t \cup \{x\})\}$$

for a random $i$. Since these items are not selected by GREEDY, adding them to $T_i^t$ will not change the solution. Therefore GREEDY$(T_i^t) = $ GREEDY$(T_i^t \cup O^t)$. Let $S_i^t = $ GREEDY$(T_i^t)$ be the solution of machine $i$ in round $t$. Since GREEDY provides an $\alpha$ approximation, and $f$ is monotone, it holds that

$$f(S_i^t) = \text{GREEDY}(T_i^t \cup O^t) \geq \alpha \cdot f(O^t).$$

We conclude that expected value of the solution of a random machine in round $t$ is at least $\alpha \cdot \mathbb{E}[f(O^t)]$. Since the final returned set $S$ has the maximum value over all intermediate solutions it follows that

$$\mathbb{E}[f(S)] \geq \alpha \cdot \mathbb{E}[f(O^t)]. \quad (4)$$

On the other hand, in the last round all items $A_r$ are collected on one machine, and GREEDY selects a solution among them. For any $x \in$ OPT, let $q_x$ be the probability that item $x$ is present in the last round (not pruned in any round). Let OPT$^S$ to be the set of these items, i.e.

$$\text{OPT}^S = \text{OPT} \cap A_r.$$

Since GREEDY is an $\alpha$ approximation, the expected value of solution in the last round and consequently the expected value $\mathbb{E}[f(S)]$ are both at least $\alpha \cdot \mathbb{E}[f(\text{OPT}^S)]$. To conclude, we exploit the properties of the Lovász extension as follows. Let $q$ and $p^t$ be vectors with $q_x$ and $p_x^t$ at component $x$, respectively. It follows that

$$\mathbb{E}[f(\text{OPT}^S)] + \sum_{t=1}^{r-1} \mathbb{E}[f(O^t)]$$
$$= \mathbb{E}[f^L(\mathbb{1}_{\text{OPT}^S})] + \sum_{t=1}^{r-1} \mathbb{E}[f^L(\mathbb{1}_{O^t})]$$
$$\geq f^L(\mathbb{E}[\mathbb{1}_{\text{OPT}^S}]) + \sum_{t=1}^{r-1} f^L(\mathbb{E}[\mathbb{1}_{O^t}])$$
$$\geq f^L(q) + \sum_{t=1}^{r-1} f^L(p^t)$$
$$\geq f^L(\mathbb{1}_{\text{OPT}}) = f(\text{OPT}).$$

The first equality is implied by the Lovász extension. It is followed by Jensen's inequality exploiting the convexity of $f^L$. The second inequality follows by definition. The last inequality is implied by the fact that each $x \in$ OPT is either omitted in one of the rounds, or is present in OPT$^S$. We note that the probability of $x$ being omitted in round $t$ is upper bounded by $p_x^t$, because it might have been omitted in one of the earlier rounds.

Therefore, by applying (4) we conclude that

$$f(\text{OPT}) \leq \mathbb{E}[f(\text{OPT}^S)] + \sum_{t=1}^{r-1} \mathbb{E}[f(O^t)]$$
$$\leq \frac{1}{\alpha}\mathbb{E}[f(S)] + \frac{r-1}{\alpha}\mathbb{E}[f(S)]$$
$$\leq \frac{r}{\alpha}\mathbb{E}[f(S)]$$

which completes the proof. □

## 4. Experiments

In this section we empirically validate two main claims of the paper. We first demonstrate that the proposed algorithm scales horizontally. We then show that the available capacity does not have a significant effect on the approximation quality. We compare the performance of the proposed algorithm to the the two-phase RANDGREEDI, centralized GREEDY run on the entire data set, and a randomly selected subset of size $k$. Finally, in a set of large-scale experiments we investigate the usage of STOCHASTIC GREEDY as the compression subprocedure. The data sets and objective functions are summarized in Table 2.

### 4.1. Data Sets

CSN. The Community Seismic Network uses smart phones with accelerometers as inexpensive seismometers for earthquake detection. In Faulkner et al. (2011), 7 GB of acceleration data was recorded from volunteers carrying and operating their phone in normal conditions (walking, talking, on desk, etc.). From this data, 17-dimensional feature vectors were computed (containing frequency information, moments, etc.).

TINY IMAGES. In our experiments we used two subsets of the Tiny Images data set consisting of $32 \times 32$ RGB images, each represented as a 3072 dimensional vector (Torralba et al., 2008). We normalized the vectors to zero mean and unit norm. Following Mirzasoleiman et al. (2013), we select a fixed random subsample of 10 000 elements for evaluation on each machine.

PARKINSONS. The data set consists of 5875 biomedical voice measurements with 22 attributes from people with early-stage Parkinsons disease (Tsanas et al., 2010). We normalized the vectors to zero mean and unit norm.

YAHOO! WEBSCOPE R6A. This data set contains a fraction of the user click log for news articles displayed in the Featured Tab of the Today Module on Yahoo! Front Page during the first ten days in May 2009. The data set contains approximately 45 000 000 user visits to the Today Module. In addition to the full data set, we also considered a subset of size 100 000.

| DATA SET | N | D | $f(S)$ |
|---|---|---|---|
| PARKINSONS | 5800 | 22 | LOGDET |
| WEBSCOPE-100K | 100 000 | 6 | LOGDET |
| CSN-20K | 20 000 | 17 | EXEMPLAR |
| TINY-10K | 10 000 | 3074 | EXEMPLAR |
| TINY | 1 000 000 | 3074 | EXEMPLAR |
| WEBSCOPE | 45 000 000 | 6 | LOGDET |

Table 2. Data set size, dimension, and the objective function.

## 4.2. Objective Functions

**Exemplar-based clustering.** A classic way to select a set of exemplars that best represent a massive data set is to solve the $k$-medoid problem (Kaufman & Rousseeuw, 1987) by minimizing the sum of pairwise dissimilarities between exemplars $A$ and elements of the data set $V$. This problem can be converted to a submodular maximization problem subject to a cardinality constraint as follows: First, define $L(S) := \frac{1}{V} \min_{v \in S} d(e, v)$ where $d : V \times V \to \mathbb{R}_+$ is a distance function, encoding the dissimilarity between elements. Then, the function that measures the reduction in quantization error by using elements in $S$,

$$f(S) = L(\{e_0\}) - L(S \cup \{e_0\}),$$

is monotone submodular and maximizing $f$ is equivalent to minimizing $L$ (Krause & Golovin, 2012). We performed several exemplar-based clustering experiments with the distance function set to $d(x, y) = ||x - y||^2$ and using the zero vector as the auxiliary element $e_0$. To exactly evaluate $f$ in examplar-based clustering, one needs to have access to the full data set on each machine. Nevertheless, as this function is additively decomposable and bounded, it can be approximated to arbitrary precision by an appropriately scaled sum over a random subsample (Chernoff bound). We select a set of $k$ most representative signals using exemplar based clustering.

**Active set selection.** In nonparametric learning (e.g. Sparse Gaussian Processes) we wish to find a set of representative samples. Formally a GP is a joint probability distribution over a (possibly infinite) set of random variables $X_V$, indexed by our ground set $V$, such that every (finite) subset $X_S$ for $S = \{e_1, \ldots, e_s\}$ is distributed according to a multivariate normal distribution, i.e., $P(X_S = x_S) = N(x_S; \mu_S, \Sigma_{S,S})$. $\mu_S = (\mu_{e_1}, \ldots, \mu_{e_s})$ and $\Sigma_{S,S} = [K_{e_i,e_j}](1 \leq i, j \leq k)$ are the prior mean vector and prior covariance matrix, respectively. The covariance matrix is parametrized via a (positive definite kernel) function $K$. A commonly used kernel function is the squared exponential kernel $K(e_i, e_j) = \exp{-||e_i - e_j||2/h^2}$. Upon observations $y_A = x_A + n_A$ (where $n_A$ is a vector of independent Gaussian noise with variance $\sigma^2$), the predictive distribution of a new data point $e \in V$ is a normal distribution $P(X_e|y_A) = \mathcal{N}(\mu_{e|A}, \Sigma^2_{e|A})$, where

$$\mu_{e|A} = \mu_e + \Sigma_{e,A}(\Sigma_{A,A} + \sigma^2 I)^{-1}(x_A - \mu_A)$$

$$\sigma^2_{e|A} = \sigma^2_e - \Sigma_{e,A}(\Sigma_{A,A} + \sigma^2 I)^{-1}\Sigma_{A,e}$$

Note that evaluating the latter is computationally expensive as it requires a matrix inversion. Instead, most efficient approaches for making predictions in GPs rely on choosing a small – so called active – set of data points. For instance, in the Informative Vector Machine, one seeks a set $S$ such that the information gain $f(S) = I(Y_S; X_V) = H(X_V) - H(X_V|Y_S) = \frac{1}{2} \log \det(I + \sigma^{-2}\Sigma_{S,S})$ is maximized. This choice of $f(S)$ is monotone submodular (Krause & Golovin, 2012). We perform several experiments optimizing the active set selection objective with a Gaussian kernel ($h = 0.5$ and $\sigma = 1$).

| DATASET | K | $\mu_1$ | $\mu_2$ | $\mu_3$ | RANDOM |
|---|---|---|---|---|---|
| WEB-100K | 50 | 0.01 | 0.00 | 0.02 | 21.17 |
| WEB-100K | 100 | 0.01 | 0.01 | 0.00 | 23.06 |
| CSN | 50 | 0.05 | 0.07 | 0.04 | 51.21 |
| CSN | 100 | 0.05 | 0.05 | 0.04 | 41.40 |
| PARKINSONS | 50 | 0.36 | 0.04 | 0.14 | 31.36 |
| PARKINSONS | 100 | 0.11 | 0.06 | 0.13 | 33.38 |
| TINY-10K | 50 | 0.08 | 0.15 | 0.04 | 59.41 |
| TINY-10K | 100 | 0.34 | 0.32 | 0.13 | 47.37 |

*Table 3.* Relative error (in percentage) with respect to the centralized GREEDY for three fixed capacities $\mu$ (200, 400 and 800).

## 4.3. Impact of Capacity on the Approximation factor

In the first set of experiments, we study the objective function value as a function of the available capacity. We consider three baseline methods: random subset of $k$ elements, centralized GREEDY on the full data set, and the two-round RANDGREEDI by Barbosa et al. (2015a). We use the lazy variant of the GREEDY algorithm (Minoux, 1978) as the $\beta$-nice algorithm in our multi-round proposal. For each algorithm we report the ratio between the obtained function value and the one obtained by the centralized GREEDY averaged over 10 trials.

Figures 2 (a) and (c) show the results of active set selection with $k = 50$. Figures 2 (b) and (d) show the results of exemplar-based clustering with $k = 50$. The vertical line represents the lower-bound on the capacity required for the two-round algorithms as per Table 1. We observe that the proposed algorithm TREE achieves performance close to the centralized greedy solution, even with extremely limited capacity of $2k$. As expected, if the capacity exceeds $\sqrt{nk}$, the performance matches that of RANDGREEDI.

Table 3 shows the impact of the available capacity on the relative approximation error for various values of $k$. The last column represents the relative error of a randomly selected subset of size $k$. We remark that the algorithm achieves a relative error of less than 1% across all data sets.

## 4.4. Large-scale Experiments

In the second set of experiments, we apply both GREEDY and STOCHASTIC GREEDY (TREE and STOCHASTIC-TREE, respectively) as pruning subprocedures. We consider the full WEBSCOPE data set and a subset of 1 000 000

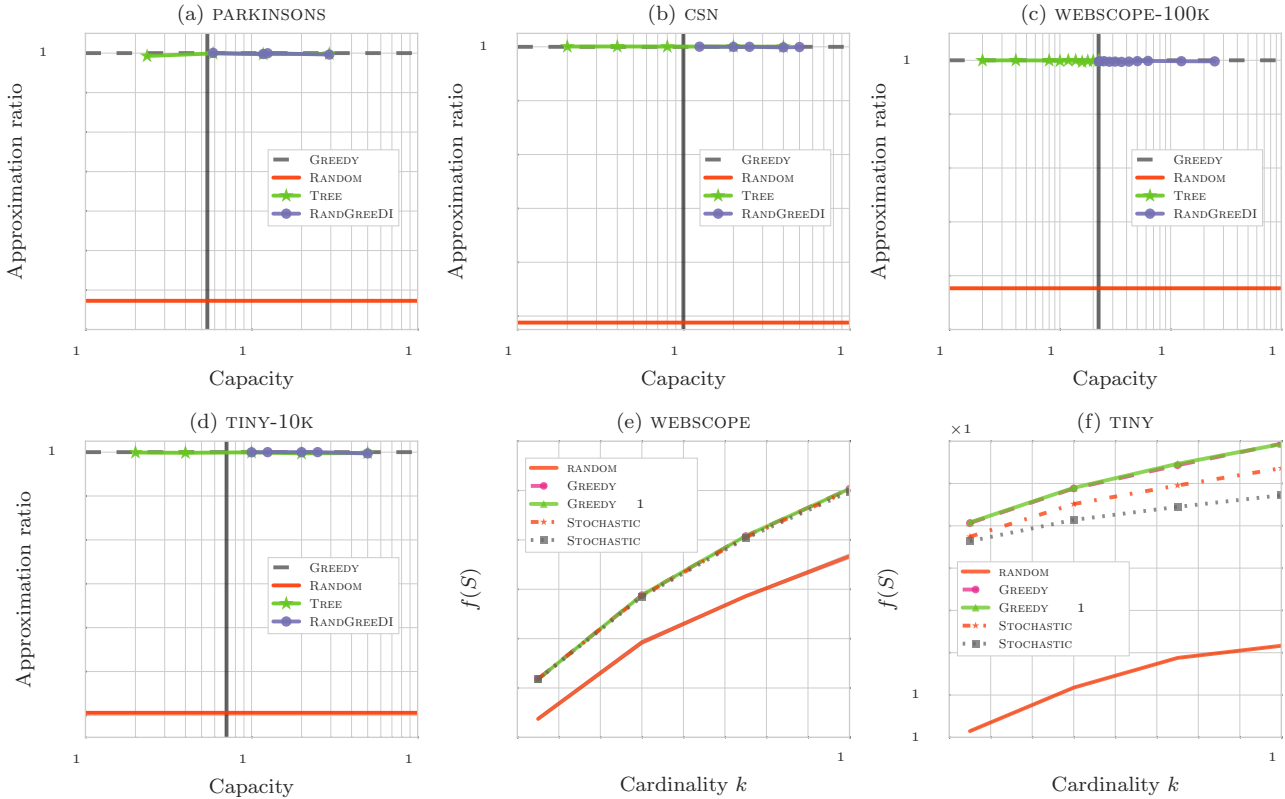

*Figure 2.* (a)-(d) show the approximation ratio with respect to the centralized GREEDY for varying capacity. The vertical gray line represents the necessary capacity for the two-round algorithms. We observe that the TREE algorithm is able to cope with extremely limited capacity (e.g. $2k$). Figures (e) and (f) show the results of the large-scale experiments using both GREEDY and STOCHASTIC GREEDY as pruning subprocedures. A slight decrease in approximation quality due to stochastic optimization is visible in (f).

TINY IMAGES. The capacity is set to a small percentage of the ground set size ($0.05\%$ and $0.1\%$). Furthermore, we consider two instances of STOCHASTIC GREEDY, one with $\varepsilon = 0.5$ and the other with $\varepsilon = 0.2$, both with capacity of $0.05\%$ of the ground set size.

Figure 2(e) shows the results of active set selection on WEBSCOPE. We observe that both versions of the proposed algorithm match the performance of centralized GREEDY, even when the available capacity is extremely limited. Figure 2(f) shows the results of exemplar-based clustering on TINY. We observe a slight loss in the approximation quality when using STOCHASTIC GREEDY. Our intuition is that the variance introduced by stochastic optimization is larger in the context of exemplar-based clustering.

## 5. Conclusion

Existing approaches to distributed submodular maximization rely on the implicit assumption that the capacity of each machine grows as the data set size increases. To the best of our knowledge, we present the first framework for constrained distributed submodular maximization that is *horizontally scalable*: It scales to larger problem instances by using more machines of limited capacity. Our framework is based on a multi-round approach, whereby in each round a fraction of the elements are discarded until all the remaining elements fit on one machine. We provide approximation guarantees for distributed submodular maximization under cardinality constraints and extend the results to hereditary constraints. The proposed framework adapts to the available capacity: If the capacity is larger than the data set size, it emulates the centralized GREEDY. If the capacity is at least $\sqrt{nk}$, it reduces to the existing two-round approaches. Otherwise, it proceeds in multiple rounds. We empirically evaluate the proposed approach on a variety of data sets and demonstrate that the algorithm achieves performance close to the centralized GREEDY solution, even with extremely limited capacity.

## Acknowledgements

We thank the reviewers for their insightful comments. This research was partially supported by ERC StG 307036 and the Zurich Information Security Center.

Horizontally Scalable Submodular Maximization## References

Badanidiyuru, Ashwinkumar and Vondrák, Jan. Fast algorithms for maximizing submodular functions. In *ACM-SIAM Symposium on Discrete Algorithms*, pp. 1497–1514. SIAM, 2014.

Barbosa, Rafael, Ene, Alina, Nguyen, Huy L, and Ward, Justin. The power of randomization: Distributed submodular maximization on massive datasets. In *International Conference on Machine Learning*, pp. 1236–1244, 2015a.

Barbosa, Rafael da Ponte, Ene, Alina, Nguyen, Huy L, and Ward, Justin. A new framework for distributed submodular maximization. *arXiv preprint arXiv:1507.03719*, 2015b.

Faulkner, Matthew, Olson, Michael, Chandy, Rishi, Krause, Jonathan, Chandy, K. Mani, and Krause, Andreas. The next big one: Detecting earthquakes and other rare events from community-based sensors. In *ACM/IEEE International Conference on Information Processing in Sensor Networks*, 2011.

Golovin, Daniel and Krause, Andreas. Adaptive submodularity: Theory and applications in active learning and stochastic optimization. *Journal of Artificial Intelligence Research*, pp. 427–486, 2011.

Kaufman, Leonard and Rousseeuw, Peter. Clustering by means of medoids. *Statistical Data Analysis Based on the L1-Norm and Related Methods*, pp. North–Holland, 1987.

Kempe, David, Kleinberg, Jon, and Tardos, Éva. Maximizing the spread of influence through a social network. In *ACM SIGKDD International Conference on Knowledge Discovery and Data Mining*, pp. 137–146. ACM, 2003.

Krause, Andreas and Golovin, Daniel. Submodular function maximization. *Tractability: Practical Approaches to Hard Problems*, 3:19, 2012.

Kumar, Ravi, Moseley, Benjamin, Vassilvitskii, Sergei, and Vattani, Andrea. Fast greedy algorithms in mapreduce and streaming. In *ACM Symposium on Parallelism in Algorithms and Architectures*, pp. 1–10. ACM, 2013.

Lin, Hui and Bilmes, Jeff. A class of submodular functions for document summarization. In *Annual Meeting of the Association for Computational Linguistics: Human Language Technologies-Volume 1*, pp. 510–520. Association for Computational Linguistics, 2011.

Minoux, Michel. Accelerated greedy algorithms for maximizing submodular set functions. In *Optimization Techniques*, pp. 234–243. Springer, 1978.

Mirrokni, Vahab and Zadimoghaddam, Morteza. Randomized composable core-sets for distributed submodular maximization. In *ACM Symposium on Theory of Computing*, pp. 153–162. ACM, 2015.

Mirzasoleiman, Baharan, Karbasi, Amin, Sarkar, Rik, and Krause, Andreas. Distributed submodular maximization: Identifying representative elements in massive data. In *Advances in Neural Information Processing Systems*, pp. 2049–2057, 2013.

Mirzasoleiman, Baharan, Badanidiyuru, Ashwinkumar, Karbasi, Amin, Vondrak, Jan, and Krause, Andreas. Lazier than lazy greedy. In *AAAI Conference on Artificial Intelligence*, 2015.

Nemhauser, George L, Wolsey, Laurence A, and Fisher, Marshall L. An analysis of approximations for maximizing submodular set functions-I. *Mathematical Programming*, 14(1):265–294, 1978.

Torralba, Antonio, Fergus, Rob, and Freeman, William T. 80 million tiny images: A large data set for nonparametric object and scene recognition. *Pattern Analysis and Machine Intelligence, IEEE Transactions on*, 30 (11):1958–1970, 2008.

Tsanas, Athanasios, Little, Max A, McSharry, Patrick E, and Ramig, Lorraine O. Enhanced classical dysphonia measures and sparse regression for telemonitoring of Parkinson's disease progression. In *International Conference on Acoustics, Speech and Signal Processing*, pp. 594–597, 2010.

## A. Detailed Analysis of Lemma 3.4

Let $\Delta(x, X)$ denote the marginal value of adding item $x$ to set $X$, i.e., $\Delta(x, X) \stackrel{\text{def}}{=} f(X \cup \{x\}) - f(X)$. The submodularity property for $X \subseteq Y \subseteq V$ and item $x \in V \setminus Y$ can now be expressed as $\Delta(x, X) \geq \Delta(x, Y)$. Let $\text{OPT} = \arg\max_{|S| \leq k} f(S)$, and $\text{OPT}^S$ be the set of selected items of OPT in the partial solutions

$$\text{OPT}^S \stackrel{\text{def}}{=} \text{OPT} \cap (\cup_{i=1}^m S_i).$$

**Lemma 3.4.** *Consider an arbitrary subset $B \subseteq V$, and a random partitioning of $B$ into $L$ sets $T_1, T_2, \cdots, T_L$. Let $S_i$ be the output of algorithm $\mathcal{A}$ on $T_i$. If $\mathcal{A}$ is $\beta$-nice, for any subset $C \subseteq B$ with size at most $k$, it holds that*

$$\mathbb{E}[f(C^S)] \geq f(C) - (1 + \beta)\mathbb{E}\left[\max_{1 \leq i \leq L} f(S_i)\right]$$

*where $C^S = C \cap \left(\cup_{1 \leq i \leq L} S_i\right)$.*

**Proof of Lemma 3.4** The proof builds on the result of Mirrokni & Zadimoghaddam (2015). The critical difference is that it holds for *any subset* of $C$, not only OPT. To quantify the contributions of items in $C$, we introduce the following formalization. Let $\pi$ be an arbitrary permutation on items of $C$. For each $x \in C$, define $\pi^x$ to be the set of items that appear prior to $x$ in $\pi$. Hence, $f(C) = \sum_{x \in C} \Delta(x, \pi^x)$. Let $G_i = (C \cap T_i) \setminus S_i$. We are going to show the following:

$$f(C^S) \geq f(C) - \sum_{x \in C \setminus C^S} \Delta(x, \pi^x) \tag{5}$$

$$\sum_{x \in C \setminus C^S} \Delta(x, \pi^x) = \sum_{i=1}^{L} \sum_{x \in G_i} \Delta(x, \pi^x) \leq \beta \max_{1 \leq i \leq L} f(S_i) + \sum_{i=1}^{L} \sum_{x \in G_i} (\Delta(x, \pi^x) - \Delta(x, S_i)) \tag{6}$$

$$\mathbb{E}\left[\sum_{i=1}^{L} \sum_{x \in G_i} \Delta(x, \pi^x) - \Delta(x, S_i)\right] \leq \frac{1}{L}\mathbb{E}\left[\sum_{i=1}^{L} f(S_i)\right] \leq \max_{1 \leq i \leq L} f(S_i) \tag{7}$$

Given these results, we can prove the claim as follows:

$$\begin{aligned}
\mathbb{E}[f(C^S)] &\geq f(C) - \sum_{x \in C \setminus C^S} \mathbb{E}[\Delta(x, \pi^x)] \\
&\geq f(C) - \beta\mathbb{E}\left[\max_{1 \leq i \leq L} f(S_i)\right] + \sum_{i=1}^{L} \sum_{x \in G_i} \mathbb{E}\left[\Delta(x, \pi^x) - \Delta(x, S_i)\right] \\
&\geq f(C) - \beta\mathbb{E}\left[\max_{1 \leq i \leq L} f(S_i)\right] + \max_{1 \leq i \leq L} f(S_i).
\end{aligned}$$

where the expectation is taken over the random partitioning $\{T_1, \cdots, T_L\}$. We now prove the intermediate results.

Inequality (5) follows from submodularity of $f$. In particular, by definition of $\Delta$ and $\pi^x$, it holds that $f(C) - f(C^S) = \sum_{x \in C \setminus C^S} \Delta(x, \pi^x \cup C^S)$. By submodularity, it holds that $\Delta(x, \pi^x \cup C^S) \leq \Delta(x, \pi^x)$ since $\pi^x$ is a subset of $\pi^x \cup C^S$. Therefore $f(C) - f(C^S) \leq \sum_{x \in C \setminus C^S} \Delta(x, \pi^x)$.

Inequality (6) follows from the following decomposition. Since $x \in C$ is sent to one of the $L$ machines ($x \in T_i$ for some $1 \leq i \leq L$) it holds that

$$\begin{aligned}
\sum_{x \in C \setminus C^S} \Delta(x, \pi^x) &= \sum_{1 \leq i \leq L} \sum_{x \in G_i} \Delta(x, \pi^x) \\
&= \sum_{i=1}^{L} \sum_{x \in C \cap T_i \setminus S_i} (\Delta(x, \pi^x \cup S_i) + (\Delta(x, \pi^x) - \Delta(x, \pi^x \cup S_i)))
\end{aligned}$$

By submodularity, the first term is at most $\Delta(x, S_i)$. Furthermore, since $x$ was not selected in machine $i$, we can upper-bound the first term by $\beta \frac{f(S_i)}{k}$. The result follows from $f(S_i) \leq \max_{1 \leq i' \leq L} f(S_{i'})$, and the fact that there are at most $k$ items in $C \setminus C^S = \cup_{i=1}^{L} (C \cap T_i \setminus S_i)$.

Finally, to prove Inequality (7) we use randomness of partition $\{T_1, T_2, \cdots, T_L\}$. The main idea is to show that the sum of the $\Delta$ differences in the third claim is in expectation at most $\frac{1}{m}$ fraction of sum of $\Delta$ differences for a larger set of pairs $(i, x)$. In particular, we show that

$$\mathbb{E}\left[\sum_{i=1}^{L} \sum_{x \in C \cap T_i \setminus S_i} \Delta(x, \pi^x) - \Delta(x, \pi^x \cup S_i)\right] \leq \frac{1}{L} \cdot \mathbb{E}\left[\sum_{i=1}^{L} \sum_{x \in C} \Delta(x, \pi^x) - \Delta(x, \pi^x \cup S_i)\right].$$

To simplify the rest of the proof, let $A$ be the left hand side of the above inequality, and $B$ be its right hand side. Also to simplify expressions $A$ and $B$, we introduce the following notation: For every item $x$ and set $T \subseteq B \subseteq V$, let $h(x, T)$ denote $\Delta(x, \pi^x) - \Delta(x, \pi^x \cup \mathcal{A}(T))$. Also let $\mathbb{1}[x \notin \mathcal{A}(T \cup \{x\})]$ be equal to one if $x$ is not in set $\mathcal{A}(T \cup \{x\})$, and zero otherwise. We note that $A$ and $B$ are both separable for different choices of item $x$ and set $T_i$, and can be rewritten formally using the new notation as follows:

$$A = \sum_{i=1}^{L} \sum_{x \in C} \sum_{T \subseteq V \setminus \{x\}} Pr[T_i = T \cup \{x\}] \mathbb{1}[x \notin \mathcal{A}(T \cup \{x\})] h(x, T \cup \{x\})$$

$$B = \sum_{i=1}^{L} \sum_{x \in C} \sum_{T \subseteq V \setminus \{x\}} (Pr[T_i = T \cup \{x\}] h(x, T \cup \{x\}) + Pr[T_i = T] h(x, T))$$

$$\geq \sum_{i=1}^{L} \sum_{x \in C} \sum_{T \subseteq V \setminus \{x\}} \mathbb{1}[x \notin \mathcal{A}(T \cup \{x\})] h(x, T \cup \{x\}) (Pr[T_i = T \cup \{x\}] + Pr[T_i = T])$$

where the inequality is implied by the following observations. Function $h$ is non-negative, so multiplying the sum by $\mathbb{1}[x \notin \mathcal{A}(T \cup \{x\})]$ (which is either zero or one) can only decrease its value. We can also replace one $h(x, T)$ with $h(x, T \cup \{x\})$ which does not change the value of the sum at all because when $\mathbb{1}[x \notin \mathcal{A}(T \cup \{x\})] = 1$ (its only non-zero value), $\mathcal{A}(T \cup \{x\})$ is identical to $\mathcal{A}(T)$ using the first property of $\beta$-nice algorithms, and thus $h(x, T) = h(x, T \cup \{x\})$.

Now we can compare $A$, and $B$ as follows: For any set $T \subseteq V \setminus \{x\}$, we have $Pr[T_i = T]$ and $Pr[T_i = T \cup \{x\}]$ are equal to $\left(\frac{1}{L}\right)^{|T|} \left(1 - \frac{1}{L}\right)^{|V|-|T|}$ and $\left(\frac{1}{L}\right)^{|T|+1} \left(1 - \frac{1}{L}\right)^{|V|-|T|-1}$ respectively. As a result, it holds that

$$\frac{Pr[T_i = T \cup \{x\}]}{Pr[T_i = T \cup \{x\}] + Pr[T_i = T]} = \frac{1}{L}$$

which implies that $A \leq B$. To complete the proof, it suffices to prove that $B \leq \frac{1}{L} \mathbb{E}\left[\sum_{i=1}^{L} f(S_i)\right]$. For any $i$,

$$\sum_{x \in C} \Delta(x, \pi^x \cup S_i) = f(C \cup S_i) - f(S_i)$$

and $\sum_{x \in C} \Delta(x, \pi^x) = f(C)$. As $f$ is monotone it holds that $f(\text{OPT} \cup S^i) \geq f(\text{OPT})$ which completes the proof. □